\documentclass[sigconf,nonacm]{acmart}

\AtBeginDocument{%
	}

\setcopyright{none}
\AtBeginDocument{%
	}

\setcopyright{none}
\begin{document}
	
	\title{Structured Evidence Selection for Weakly Supervised Video Anomaly Detection}
	
	\author{Chenglizhao Chen}
	\affiliation{%
		\institution{China University of Petroleum (East China)}
		\city{Qingdao}
		\country{China}
	}
	
	\author{Tianxiang Nan}
	\affiliation{%
		\institution{China University of Petroleum (East China)}
		\city{Qingdao}
		\country{China}
	}
	
	\author{Wen Li}
	\affiliation{%
		\institution{China University of Petroleum (East China)}
		\city{Qingdao}
		\country{China}
	}
	
	\author{Xinyu Liu}
	\authornote{Corresponding author.}
	\affiliation{%
		\institution{China University of Petroleum (East China)}
		\city{Qingdao}
		\country{China}
	}
	\email{liuxy001005@163.com}
	
	\author{Guisheng Zhang}
	\affiliation{%
		\institution{China University of Petroleum (East China)}
		\city{Qingdao}
		\country{China}
	}
	
	\author{Mengke Song}
	\affiliation{%
		\institution{China University of Petroleum (East China)}
		\city{Qingdao}
		\country{China}
	}
	
	\author{Xiaomin Yu}
	\affiliation{%
		\institution{The Hong Kong University of Science and Technology (Guangzhou)}
		\city{Guangzhou}
		\country{China}
	}
	
	\renewcommand{\shortauthors}{Chen et al.}
	
	\begin{abstract}
		Weakly supervised video anomaly detection relies solely on video-level labels for training, making it difficult to accurately localize anomalous events in complex scenes. In real-world videos, anomalous behaviors exhibit large variations in appearance and temporal duration, while scene appearance and action dynamics are often tightly entangled. Consequently, existing models tend to rely on scene-related statistical cues rather than true behavioral deviations, resulting in unstable detection performance.
		To address this challenge, we propose a Structured Evidence Selection framework (SESAD) that reformulates anomaly detection as a structured reasoning process over clip-level visual evidence. Instead of directly mapping aggregated features to anomaly scores, SESAD reorganizes clip representations into semantically structured candidate evidence and performs context-conditioned selection under scene and action constraints. This mechanism adaptively emphasizes anomaly-relevant semantics while suppressing scene interference, thereby alleviating semantic entanglement under weak supervision. Furthermore, we introduce a lightweight geometric discrimination module that constructs a dual-prototype structure in the embedding space, enabling anomaly decisions through relative geometric relations.
		Extensive experiments on UBnormal, ShanghaiTech, and UCF-Crime show that SESAD achieves 67.92, 97.99, and 88.46 AUC, respectively, while maintaining high computational efficiency and overall consistently stable anomaly discrimination.
		
	\end{abstract}

	
	\maketitle

\section{Introduction}

Video anomaly detection aims to identify events that deviate from normal behavior patterns in complex real-world environments~\cite{liu2025networking,kim2025unsupervised,liu2024generalized}. It plays an important role in intelligent surveillance, public safety, and video understanding~\cite{liu2025privacy}. In practice, anomalous events are rare and diverse, making precise temporal annotations very costly~\cite{pazho2025towards}. To reduce the annotation burden, weakly supervised video anomaly detection (WS-VAD) has attracted increasing attention. Under this setting, models are trained using only video-level labels without clip-level annotations, which significantly increases the difficulty of reliable anomaly localization~\cite{yang2026attention}.

\begin{figure}[!t]
	\centering{\includegraphics[width=1\linewidth]{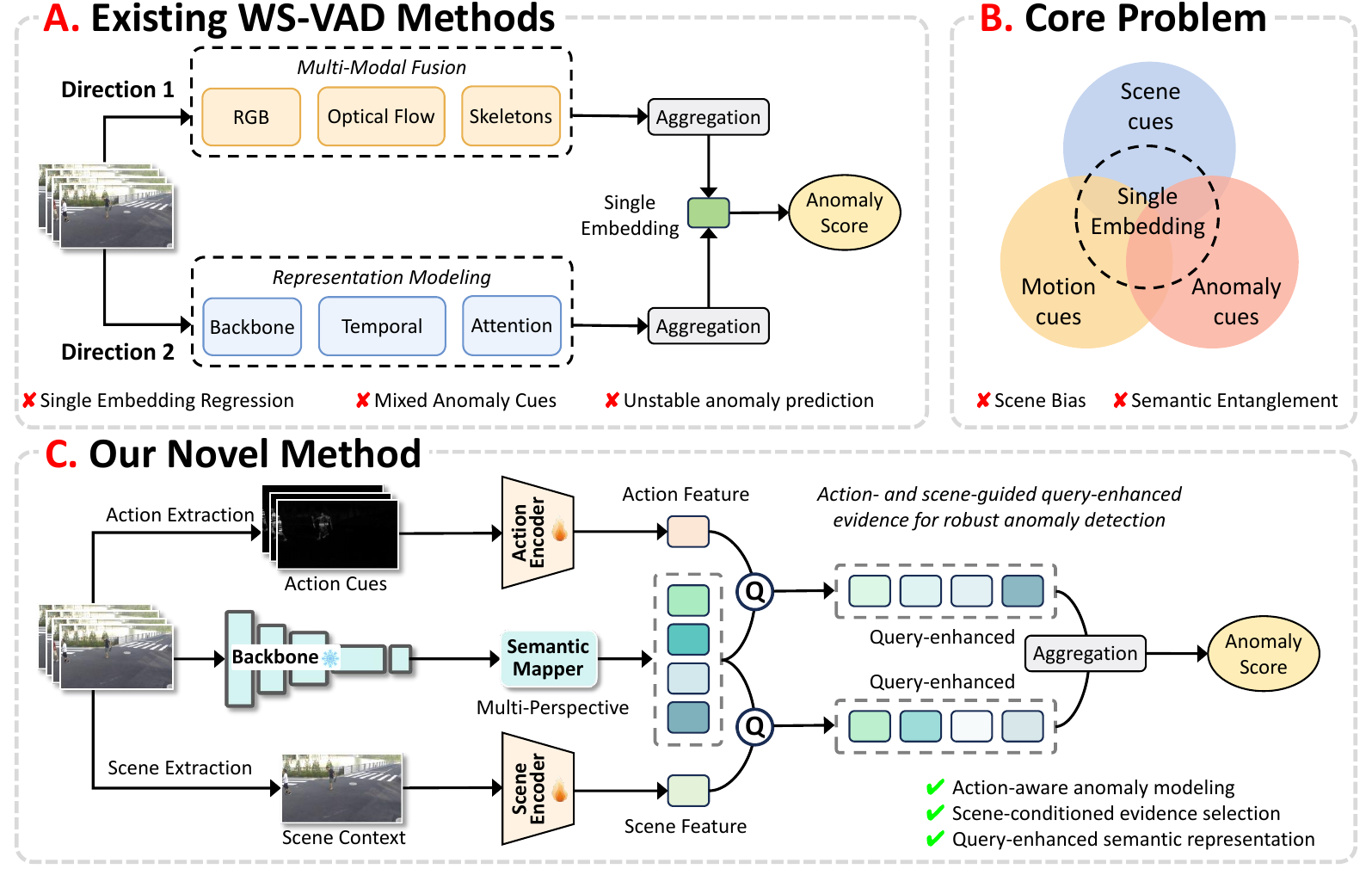}}
	\caption{Motivation and overview of our method. Existing WS-VAD methods compress heterogeneous cues into a single embedding for anomaly regression, leading to scene bias and semantic entanglement. Our approach instead performs structured evidence selection guided by scene and action queries to achieve more robust anomaly detection.}
	\label{fig:motivation}
\end{figure}

As illustrated in Figure~\ref{fig:motivation}(A), existing weakly supervised video anomaly detection methods can be broadly categorized into two research directions. The first direction introduces additional modalities, such as optical flow and skeleton information, and leverages multimodal fusion to enhance motion cues or structural information, thereby improving the discriminability of anomalous behaviors~\cite{sun2026enhancing,lyu2026bidirectional,su2025semantic,fan2024weakly}. The second direction focuses on representation modeling within RGB videos, aiming to improve feature expressiveness by designing stronger backbone networks, temporal modeling modules, or attention mechanisms~\cite{he2026mg,zhang2025dynamic,zhou2024batchnorm,pi2024fe}.
Despite differences in network architectures and information sources, these approaches share a highly consistent modeling paradigm. Specifically, videos are typically divided into multiple clips, from which clip-level features are first extracted and then aggregated into a single embedding representation. An anomaly score is subsequently regressed from this embedding to perform anomaly detection.

This single-embedding regression paradigm suffers from fundamental limitations under weak supervision. In real-world videos, scene appearance, motion dynamics, and potential anomaly cues are often highly coupled~\cite{yu2026modality,sun2023hierarchical,zhang2023exploiting}. When these heterogeneous signals are compressed into a single representation space, the model tends to rely on dominant scene statistics rather than the anomaly evidence that truly reflects behavioral deviations. As illustrated in Figure~\ref{fig:motivation}(B), the resulting embedding often mixes scene cues, motion patterns, and anomaly evidence within the same representation space, lacking an explicit mechanism for semantic disentanglement. This leads to two critical issues: scene bias, where predictions are heavily influenced by background appearance~\cite{lv2023unbiased,li2022scale}, and semantic entanglement, where anomaly-related signals are mixed with irrelevant visual patterns~\cite{cho2023look,wu2022self}. These issues ultimately result in unstable anomaly predictions and significantly weaken the model’s generalization ability across different scenes.

To address these issues, we revisit weakly supervised video anomaly detection through structured evidence reasoning. Rather than directly predicting anomaly scores from aggregated representations, we construct and select anomaly-related evidence before the final decision. We propose the Structured Evidence Selection framework for Anomaly Detection (SESAD), shown in Figure~\ref{fig:motivation}(C). SESAD first reorganizes clip-level features into multi-perspective semantic representations to preserve diverse anomaly cues. It then performs context-conditioned evidence selection, using scene and action queries to emphasize anomaly-relevant semantics and suppress misleading responses. Finally, a lightweight dual-prototype geometry determines anomalies by comparing clip representations with normal and abnormal prototypes in the embedding space.

This structured reasoning process enables the model to separate anomaly-related evidence from entangled visual signals and produce more stable and reliable anomaly predictions under weak supervision in complex scenarios. Extensive experiments on UBnormal, ShanghaiTech,  and UCF-Crime show that the proposed framework achieves 67.92, 97.99, and 88.46 AUC, respectively, while maintaining high computational efficiency and detection stability. The main contributions of this work are summarized as follows:

\begin{itemize}
	\item We revisit weakly supervised video anomaly detection from the perspective of structured evidence reasoning, moving beyond the conventional single-embedding regression paradigm. The framework explicitly constructs and selects anomaly-related evidence before decision making, enabling stable anomaly discrimination under weak supervision.
	
	\item We propose a structured representation mechanism that reorganizes clip-level features into multi-perspective semantic evidence and performs context-conditioned selection guided by scene and action queries, preserving diverse anomaly cues while suppressing scene bias.
	
	\item We introduce a lightweight dual-prototype geometric decision module that models anomaly detection through relative geometric relations between clip representations and normal and abnormal prototypes in the embedding space, improving structural consistency and detection stability.
\end{itemize}

\section{Related Works}
\subsection{Weakly Supervised VAD}
Weakly supervised video anomaly detection (WS-VAD) has attracted increasing attention due to the high cost of temporal annotations. In this setting, models are trained with only video-level labels without precise temporal supervision, making reliable anomaly localization particularly challenging. Early methods commonly adopt a multiple instance learning (MIL) framework, where a video is treated as a bag of clips and anomaly scores are optimized using video-level labels~\cite{sultani2018real,tian2021weakly,shao2023video,kamoona2023multiple}. Many works further improve temporal modeling through ranking losses, margin constraints, or attention-based aggregation strategies~\cite{yin2025learning,zhao2025mstagent,zhang2025dual,su2024vpe}.

Recent approaches further enhance representation learning by adopting stronger backbone networks or incorporating additional modalities such as optical flow or skeleton information to capture richer motion cues~\cite{meng2025audio,basak2024diffusion,hussain2024tds}. Despite architectural differences, most existing methods follow a similar paradigm: clip features are aggregated into a single embedding and mapped to an anomaly score~\cite{luo2025fadmb,sun2025delving}. Under weak supervision, this paradigm often mixes anomaly cues with scene-related statistics, leading to unstable predictions in complex and diverse environments~\cite{chen2025unveiling,pu2024learning}.

\subsection{Multi-Semantic Representation Learning}
Several studies attempt to represent video content from multiple semantic perspectives to capture diverse and complex behavior patterns~\cite{yu2025unicorn,qiu2025learning,chen2025dctformer}. Multi-branch or multi-view architectures are commonly adopted to separate different types of visual cues, such as appearance, motion, and temporal dynamics~\cite{wu2025flow,qiu2024video,tao2024feature}. By organizing features into different semantic subspaces, these methods aim to preserve richer contextual information and improve the overall representation capacity.

Although multi-semantic representations provide more expressive feature spaces, most existing approaches eventually aggregate these representations into a single embedding before performing anomaly prediction~\cite{cai2026semantic,sun2024dual}. As a result, anomaly-related signals from different semantic dimensions remain compressed into one representation, which limits the model’s ability to explicitly distinguish anomaly evidence from other visual patterns~\cite{leng2025dual}.   

\subsection{Context-Aware Anomaly Modeling}
Context-aware approaches incorporate contextual information to improve anomaly detection~\cite{wu2024weakly,wu2024vadclip,wu2024open}. Some methods model normal patterns or temporal consistency to detect deviations from regular events~\cite{biswas2025mmvad,dong2025diagnosis}, while others introduce scene context to reduce false positives from background variations~\cite{wang2026video,hu2025efficient}. Although contextual cues offer useful references for anomaly detection, existing approaches typically apply contextual modeling at the aggregation or score prediction stage~\cite{guo2025aligning,cai2025hiprobe,li2024multiscale,liu2024vadiffusion}. Few works explicitly construct a structured evidence space or perform context-conditioned evidence selection before the final decision process~\cite{liu2026watching}.

In contrast, our method focuses on structured evidence reasoning under weak supervision. By organizing clip-level features into multi-perspective evidence representations and performing context-conditioned evidence selection, the proposed framework enables more reliable anomaly discrimination in complex scenes.

\begin{figure*}[!t]
	\centering{\includegraphics[width=1\linewidth]{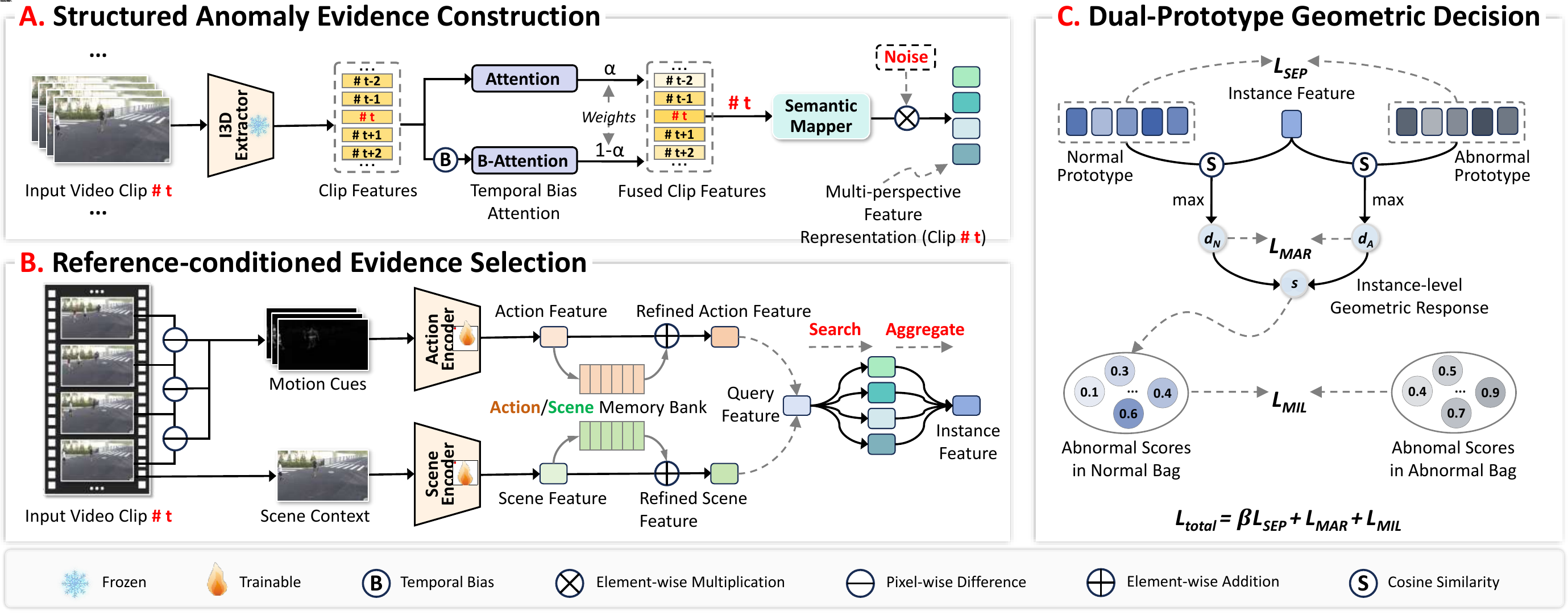}}
	\caption{Overview of the proposed framework. (A) Structured anomaly evidence construction reorganizes clip-level features into multi-perspective semantic representations. (B) Reference-conditioned evidence selection refines anomaly evidence using scene and action queries. (C) Dual-prototype geometric decision performs anomaly discrimination in the embedding space.}
	\label{fig:pipeline}
\end{figure*}

\section{Methodology}
\subsection{Overview}
Given an input video, we first divide it into a sequence of clips and extract clip-level spatiotemporal features using a pretrained I3D network~\cite{carreira2017quo} as the backbone feature extractor. Instead of directly aggregating these features into a single embedding for anomaly score regression, we formulate anomaly detection as a structured evidence reasoning process over clip-level representations.

As illustrated in Figure~\ref{fig:pipeline}, the proposed framework consists of three stages. First, the structured anomaly evidence construction module reorganizes clip-level features into multi-perspective semantic representations. Temporal relations between clips are modeled using an attention mechanism with temporal bias. The resulting features are then projected into semantic subspaces through a semantic mapping module. This step preserves diverse anomaly cues and reduces the semantic compression caused by direct feature aggregation. Second, the reference-conditioned evidence selection module introduces contextual guidance to refine anomaly evidence. Action queries and scene queries are extracted from motion cues and scene context. They are further refined through a memory-based aggregation process. The refined queries are used to retrieve and combine relevant semantic representations, producing context-aware instance features for anomaly reasoning. Finally, the dual-prototype geometric decision module models anomaly detection as a geometric relation in the embedding space. Two groups of prototypes are learned to represent normal and abnormal behavior patterns. The anomaly property of each instance is determined by its relative position with respect to these prototype sets.

\subsection{Structured Anomaly Evidence Construction}

Clip-level features extracted from videos often contain heterogeneous information, including scene appearance, normal activities, and potential anomaly cues. Under weak supervision, directly aggregating these features tends to compress mixed signals into a single representation, which may weaken anomaly-related evidence and lead to unstable anomaly prediction.

To address this issue, we construct structured anomaly evidence from clip-level features. As illustrated in Fig.~\ref{fig:pipeline}(A), the input video is first divided into clips and processed by a pretrained I3D network to extract spatiotemporal features. Given a clip sequence of length $N$, the extracted features are denoted as $X=\{x_1,x_2,\ldots,x_N\}$, where $x_i\in\mathbb{R}^D$ represents the feature of the $i$-th clip.

\noindent\textbf{Temporal Bias Attention.}
Anomalous patterns in videos may appear at different temporal ranges in real-world scenarios. 
Some anomalies manifest as long-term semantic shifts across multiple clips, while others are reflected by subtle short-range inconsistencies between neighboring clips. 
Modeling temporal relations using only a single attention scale may mix these patterns within a shared representation space. 
To address this issue, we introduce a temporal bias attention mechanism to capture both global and local temporal relations more effectively and robustly.

Given the clip-level feature sequence $X$, we first apply standard multi-head self-attention to model global dependencies among clips, producing a global relational representation $H^{g}\in\mathbb{R}^{N\times D}$.

To further introduce local temporal structure, we construct a distance-aware bias between clips. Since clips are arranged in temporal order, the row index of $X$ naturally reflects their relative temporal positions in the sequence. For the $h$-th attention head, we define the bias between clips $i$ and $j$ as follows:
\begin{equation}
	B^{(h)}_{ij} = \exp\left(-\frac{|i-j|}{\sigma_h}\right),
\end{equation}
where $\sigma_h > 0$ is a learnable scale parameter controlling the decay range. Clips that are closer in time receive larger bias weights, encouraging the model to focus on local temporal relations.

The bias is incorporated into the attention logits as:
\begin{equation}
	A^{(h)} =
	\mathrm{Softmax}
	\left(
	\frac{Q^{(h)}(K^{(h)})^\top}{\sqrt{D_h}}
	+
	\log B^{(h)}
	\right),
\end{equation}
where $D_h$ denotes the dimensionality of each attention head.

The outputs of all heads are concatenated to obtain the temporal bias representation $H^{b}\in\mathbb{R}^{N\times D}$. 

Finally, we fuse the global and bias-aware representations to produce the final clip representation:
\begin{equation}
	H = \alpha H^{g} + (1-\alpha) H^{b},
\end{equation}
where $\alpha \in (0,1)$ is a learnable fusion coefficient. The resulting feature matrix $H\in\mathbb{R}^{N\times D}$ captures both long-range semantic dependencies and short-range temporal structures, and serves as the input to the semantic mapping module.

\begin{figure}[!t]
	\centering{\includegraphics[width=1\linewidth]{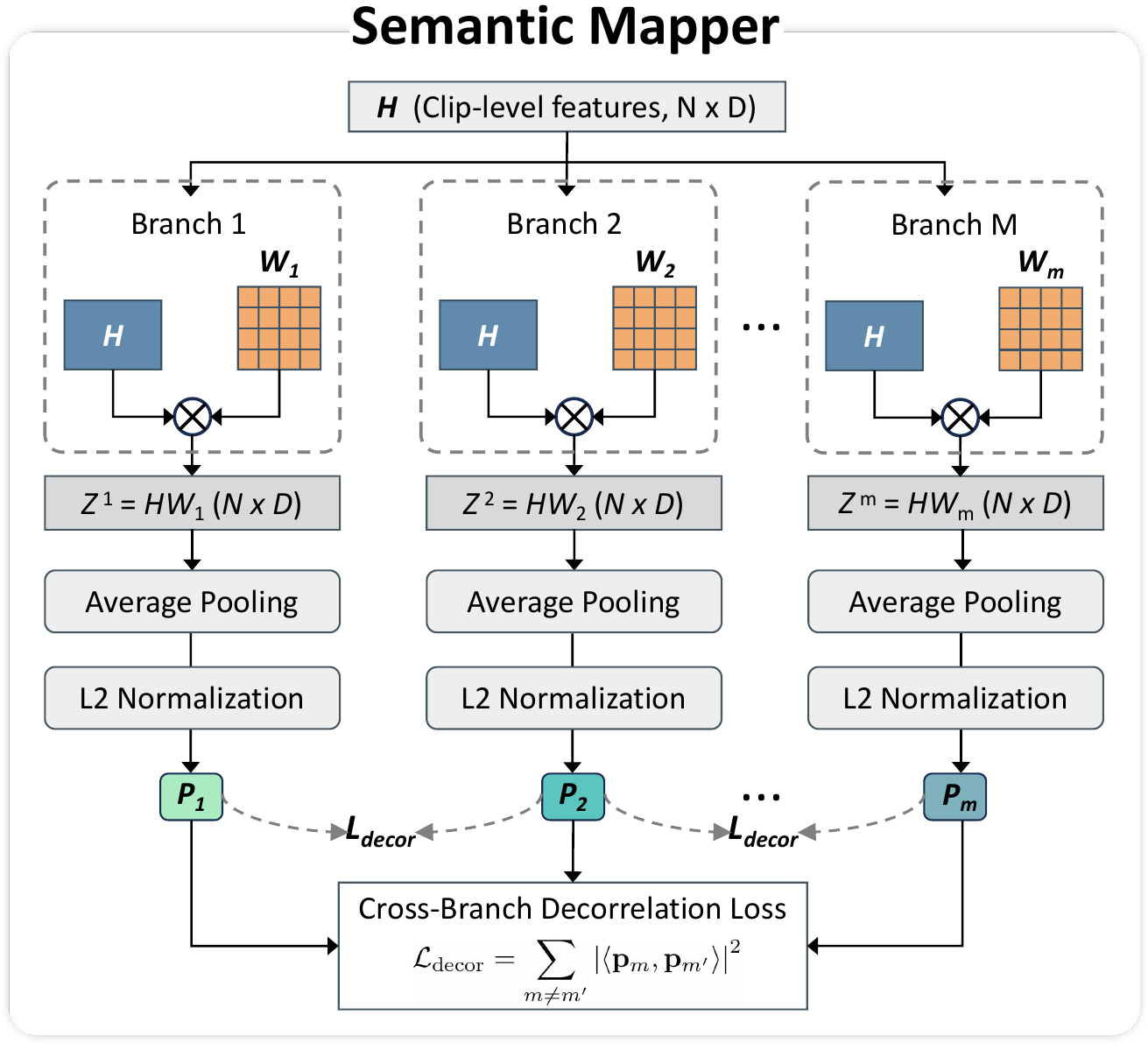}}
	\caption{Illustration of semantic mapper. Clip-level features are projected into multiple semantic branches, and a decorrelation loss encourages diverse semantic representations.}
	\label{fig:SM}
\end{figure}

\noindent\textbf{Multi-Perspective Semantic Mapping.}
After temporal relation modeling, we further transform clip features into multi-perspective semantic representations. As shown in Fig.~\ref{fig:SM}, the semantic mapper contains multiple parallel branches. Each branch projects the clip representation into a different semantic subspace, enabling the model to capture diverse anomaly cues from multiple perspectives.

Given the temporally enhanced feature matrix $H \in \mathbb{R}^{N\times D}$, the $m$-th semantic branch performs a linear projection:
\begin{equation}
	Z^{(m)} = H W_m ,
\end{equation}
where $W_m \in \mathbb{R}^{D\times D}$ denotes the projection matrix of the $m$-th branch. 
Unlike conventional aggregation-based designs, we retain the clip dimension and treat $Z^{(m)}\in\mathbb{R}^{N\times D}$ as branch-wise clip representations, where the $b$-th row $Z^{(m)}_b$ corresponds to the semantic feature of the $b$-th clip in the $m$-th semantic subspace. 
Each clip feature is further normalized as:
\begin{equation}
	\bar Z^{(m)}_b =
	\frac{Z^{(m)}_b}{\|Z^{(m)}_b\|_2}.
\end{equation}

Different semantic branches may capture similar patterns, which reduces representation diversity. To encourage each branch to focus on complementary anomaly cues, we introduce a cross-branch decorrelation constraint. 
Specifically, we first compute a branch-level summary by averaging clip features:
\begin{equation}
	p_m = \frac{1}{N}\sum_{b=1}^{N} \bar Z^{(m)}_b ,
\end{equation}
and impose a decorrelation loss:
\begin{equation}
	L_{\text{decor}}
	=
	\sum_{m\ne m'}
	\|
	p_m^\top p_{m'}
	\|^2 .
\end{equation}

This constraint penalizes correlation between semantic branches and encourages them to capture complementary anomaly cues.

The resulting clip-wise features $\{\bar Z^{(1)},\bar Z^{(2)},\ldots,\bar Z^{(M)}\}$ provide multi-perspective evidence for subsequent reasoning.

\noindent\textbf{Feature Perturbation.}
To improve representation robustness, we introduce stochastic perturbation during training. Gaussian noise is added to clip-wise features after the semantic mapping stage:
\begin{equation}
	\tilde Z^{(m)}_b = \bar Z^{(m)}_b + \epsilon ,
\end{equation}
where $\epsilon \sim \mathcal{N}(0,\sigma^2 I)$ denotes Gaussian noise with variance $\sigma^2$. 
The perturbation is applied during training to encourage stable representations and reduce overfitting, and is removed during inference.

\subsection{Reference-conditioned Evidence Selection}
After constructing multi-perspective semantic representations, the model obtains branch-wise clip features $\{\tilde Z^{(1)},\tilde Z^{(2)},\ldots,\tilde Z^{(M)}\}$, where each $\tilde Z^{(m)}\in\mathbb{R}^{N\times D}$ contains clip-wise features from the $m$-th branch. Under weak supervision, however, not all semantic responses contribute equally to anomaly recognition. Some branches may capture irrelevant scene statistics or noisy patterns. Directly aggregating these representations may therefore introduce misleading responses and weaken anomaly prediction stability.

To address this issue, we introduce a reference-conditioned evidence selection mechanism. 
As illustrated in Fig.~\ref{fig:pipeline}(B), the model uses scene and action contexts as references to guide the selection of anomaly evidence. 
These contextual references are first encoded into query representations. 
The queries are then refined through a memory bank and used to retrieve relevant semantic evidence.

\noindent\textbf{Scene and Action Queries.}
To introduce contextual references into anomaly reasoning, we construct two types of queries: 
an action query $q_a \in \mathbb{R}^D$ and a scene query $q_s \in \mathbb{R}^D$. 
These queries describe motion dynamics and scene context respectively. 
Rather than directly predicting anomaly scores, they serve as contextual conditions that modulate the importance of different semantic branches.

The action query models temporal motion patterns inside the clip. 
Since anomalous behaviors often appear as irregular motion changes, we suppress static appearance information using frame differences and extract motion-aware features:
\begin{equation}
	q_a = \mathrm{Pool}\big(\Phi_a(\mathrm{Norm}(V_{t+1}-V_t))\big) \in \mathbb{R}^D ,
\end{equation}
where $\Phi_a(\cdot)$ denotes a learnable action encoder that extracts motion-aware representations from frame differences.

The scene query captures the semantic environment in which the action occurs. 
Since scene context within a short clip is usually stable, we select a representative frame $I_c$ and extract high-level spatial features through a scene encoder $\Phi_s(\cdot)$ :
\begin{equation}
	q_s = \mathrm{Pool}\big(\Phi_s(I_c)) \in \mathbb{R}^D .
\end{equation}

\noindent\textbf{Memory-Enhanced Query Refinement.}
Directly computed queries may be sensitive to noise in video clips. 
To stabilize contextual representations, we introduce an Action/Scene Memory Bank as shown in Fig.~\ref{fig:pipeline}(B). 
The memory bank stores representative contextual patterns learned during training.

Given a query $q$, the refined query is obtained by retrieving related memory slots:
\begin{equation}
	\tilde q = \mathrm{Softmax}(qM^\top)M ,
\end{equation}
where $M$ denotes the memory matrix.

The refined action and scene queries are then fused to obtain a unified contextual reference:
\begin{equation}
	q_c = \frac{\tilde q_a + \tilde q_s}{2}.
\end{equation}

\noindent\textbf{Evidence Search and Aggregation.}
The contextual reference is used to guide the selection of anomaly evidence from the multi-perspective semantic representations. 
For the $b$-th clip, we measure the relevance between the contextual reference and the semantic feature of each branch:
\begin{equation}
	w_{b,m} = \mathrm{Softmax}_m\!\left((\tilde Z^{(m)}_b)^\top q_c\right).
\end{equation}

These weights indicate the importance of different semantic branches under the current context. 
The context-aware representation of the $b$-th clip is obtained through weighted aggregation:
\begin{equation}
	z_b = \sum_{m=1}^{M} w_{b,m}\tilde Z^{(m)}_b .
\end{equation}

This process emphasizes anomaly-relevant semantic responses while suppressing irrelevant ones. 
The resulting instance feature $\{z_b\}_{b=1}^{N}$ will be used in the subsequent geometric decision stage.

\subsection{Dual-Prototype Geometric Decision}

After reference-conditioned evidence selection, the model obtains a set of context-aware instance features 
$\{z_b\}_{b=1}^{N}$, where each $z_b \in \mathbb{R}^D$ corresponds to the representation of the $b$-th clip. 
As illustrated in Fig.~\ref{fig:pipeline}(C), instead of directly regressing an anomaly score from these features, 
we formulate anomaly detection as a geometric discrimination problem in the embedding space. 
Specifically, we construct two prototype sets representing normal and abnormal behavior patterns, 
and determine the anomaly property of each instance according to its relative geometric relation to these prototype sets.

\noindent\textbf{Dual-Prototype Geometric Space.}
Let $P_N=\{p_{N,i}\}_{i=1}^{K_N}$ and $P_A=\{p_{A,j}\}_{j=1}^{K_A}$ denote the sets of learnable normal and abnormal prototypes, respectively, where each prototype lies in $\mathbb{R}^D$. 
To remove the influence of feature magnitude and ensure comparable distances, 
all vectors are normalized onto the unit hypersphere:
\begin{equation}
	\bar{z}_b=\frac{z_b}{\|z_b\|_2}, \qquad
	\bar{p}_{N,i}=\frac{p_{N,i}}{\|p_{N,i}\|_2}, \qquad
	\bar{p}_{A,j}=\frac{p_{A,j}}{\|p_{A,j}\|_2}.
\end{equation}

Given an instance representation $\bar{z}_b$, its geometric distance to each prototype set is defined as the minimum Euclidean distance to the prototypes within the corresponding set:
\begin{equation}
	d_N(\bar{z}_b)=\min_i \|\bar{z}_b-\bar{p}_{N,i}\|_2, \qquad
	d_A(\bar{z}_b)=\min_j \|\bar{z}_b-\bar{p}_{A,j}\|_2 .
\end{equation}

In this dual-prototype-set space, anomaly is determined by the relative distance between an instance and the two semantic prototype sets. 
The two prototype groups therefore form a discriminative geometric structure that separates normal and abnormal behaviors.

\noindent\textbf{Set-level Geometric Decision.}
Based on the above geometric structure, we define the instance-level anomaly response as:
\begin{equation}
	s_b = d_N(\bar{z}_b) - d_A(\bar{z}_b),
\end{equation}
where a larger response indicates that the instance is closer to the abnormal prototype set and farther from the normal prototype set.

Under weak supervision, only video-level labels are available. Let $y\in\{0,1\}$ denote the video label, where $y=1$ indicates an anomalous video. Under the multiple instance learning (MIL) assumption, an anomalous video contains at least one abnormal instance. The video-level anomaly score is defined as:
\begin{equation}
	s_v = \max_{b} s_b .
\end{equation}

The MIL loss is formulated as a binary cross-entropy objective:
\begin{equation}
	\mathcal{L}_{MIL}
	=
	- y \log \sigma(s_v)
	- (1-y) \log \big(1-\sigma(s_v)\big),
\end{equation}
which encourages at least one instance in anomalous videos to move toward the abnormal prototype direction while suppressing the anomaly response in normal videos.

However, MIL supervision alone does not guarantee a stable embedding structure. 
To explicitly shape the dual-prototype geometry, we introduce a geometric margin constraint. 
For normal videos, instances are encouraged to stay close to the normal prototype set. 
For anomalous videos, instances are required to remain associated with at least one prototype set rather than drifting away from both.

With a fixed margin $\delta = 1$, the margin loss is defined as:
\begin{equation}
	\mathcal{L}_{MAR}
	=
	\frac{1-y}{N}\sum_{b=1}^{N}[d_N(\bar{z}_b)-1]_+
	+
	\frac{y}{N}\sum_{b=1}^{N}[\min\{d_N(\bar{z}_b),d_A(\bar{z}_b)\}-1]_+ ,
\end{equation}
where $[x]_+=\max(0,x)$.

To further prevent prototype collapse between the two semantic groups, 
we enforce a minimum separation between the normal and abnormal prototype sets:
\begin{equation}
	\mathcal{L}_{SEP}
	=
	\frac{1}{K_N K_A}
	\sum_{i=1}^{K_N}\sum_{j=1}^{K_A}
	[1-\|\bar{p}_{N,i}-\bar{p}_{A,j}\|_2]_+ .
\end{equation}

The final optimization objective is:
\begin{equation}
	\mathcal{L}_{total}
	=
	\mathcal{L}_{MIL}
	+
    \mathcal{L}_{MAR}
	+
	\beta \mathcal{L}_{SEP}.
\end{equation}

By jointly optimizing video-level supervision and geometric constraints, the embedding space forms a stable discriminative structure. Normal instances cluster around the normal prototype set, while anomalous instances move toward the abnormal prototype set. A separation term controlled by $\beta$ maintains distance between prototype groups to prevent collapse. Consequently, anomaly detection transforms from score regression into a geometric decision based on prototype relations.\textit{See Appendix C and D for details.}

\section{Experiment}
\subsection{Experimental Setup}
\noindent\textbf{Dataset.}
Experiments are conducted on three widely used benchmarks for weakly supervised video anomaly detection: UBnormal~\cite{acsintoae2022ubnormal}, ShanghaiTech~\cite{liu2018future}, and UCF-Crime~\cite{sultani2018real}. 
UBnormal is an open-set anomaly detection dataset containing 543 surveillance videos, where anomaly categories in the training and testing sets are disjoint, making the task more challenging and requiring stronger generalization ability.
ShanghaiTech consists of 437 videos captured across 13 scenes and is commonly used to evaluate anomaly localization in relatively stable surveillance environments.
UCF-Crime contains 1,900 long untrimmed surveillance videos covering 13 anomaly categories, where only video-level labels are available during training and frame-level annotations are used for evaluation.

\noindent\textbf{Evaluation Metric.}
Following standard evaluation protocols in video anomaly detection, we adopt frame-level Area Under the ROC Curve (AUC) as the primary metric, measuring the ranking consistency between predicted anomaly scores and frame-level labels. For UCF-Crime and ShanghaiTech, AUC is computed following the official protocols. For UBnormal, we report Micro-AUC computed over all testing frames. We also report Average Precision (AP) from the Precision–Recall curve as a complementary metric under the highly imbalanced distribution of normal and abnormal frames.

\noindent\textbf{Implementation.}
All experiments are conducted on a server with an NVIDIA RTX 5090 GPU. Following prior works, video features are extracted using a pretrained I3D network~\cite{carreira2017quo} trained on the Kinetics-400 dataset~\cite{carreira2017quo}, and 2048-dimensional I3D-RGB representations are obtained using the 10-crop strategy. The model is optimized using Adam with an initial learning rate of $1\times10^{-3}$ and a weight decay of $5\times10^{-3}$. Unless otherwise specified, each video is divided into $N=32$ snippets, and each snippet is represented by a $D=2048$ dimensional feature vector.\textit{See Appendix A for details}

\subsection{Comparisons with State-Of-The-Arts}
\noindent\textbf{Quantitative results.}
Table~\ref{tab:tab1} reports the quantitative comparison between our method and several representative weakly supervised video anomaly detection approaches, including DeepMIL~\cite{sultani2018real}, RTFM~\cite{tian2021weakly}, MGFN~\cite{chen2023mgfn}, UR-DMU~\cite{zhou2023dual}, RTFM-BERT~\cite{tan2024overlooked}, VadCLIP~\cite{wu2024vadclip}, OPVAD~\cite{wu2024open}, STPrompt~\cite{wu2024weakly}, MPDFL~\cite{xiao2025multilingual}, PLOVAD~\cite{xu2025plovad}, IFS-VAD~\cite{zhong2024inter}, Fed-WSVAD~\cite{wang2025federated}, and HVLMCLD~\cite{al2026hierarchical}.Following prior works, we compare with methods built upon single visual feature representations, including C3D, I3D, and CLIP features. 
This setting ensures a fair comparison by focusing on RGB-based visual representations without introducing additional modalities such as optical flow or skeleton information.

The proposed method achieves strong and consistent performance across all three benchmarks while maintaining a lightweight model size. On UBnormal, our method obtains 67.92 AUC and 73.65 AP, outperforming most previous approaches including RTFM, UR-DMU, and Fed-WSVAD, and surpassing recent CLIP-based methods in AUC. On ShanghaiTech, our approach achieves 97.99 AUC and 60.23 AP, delivering the best overall performance among the compared methods and improving upon strong baselines such as RTFM-BERT and IFS-VAD. On UCF-Crime, our method achieves 88.46 AUC and 31.21 AP, outperforming most existing methods including UR-DMU, IFS-VAD, and VadCLIP. Notably, these improvements are achieved with only 9.4MB parameters, demonstrating that structured evidence selection and geometric decision can effectively enhance anomaly discrimination while maintaining a lightweight model design. \textit{See Appendix B for details.}

\begin{table}[!t]
	\centering
	\caption{Performance comparison of different methods on UBnormal, ShanghaiTech, and UCF-Crime in terms of AUC and AP. The best results are highlighted in bold.}
	\begin{tabular}{c}
		\hspace{-0.3cm}
		\includegraphics[width=1\linewidth]{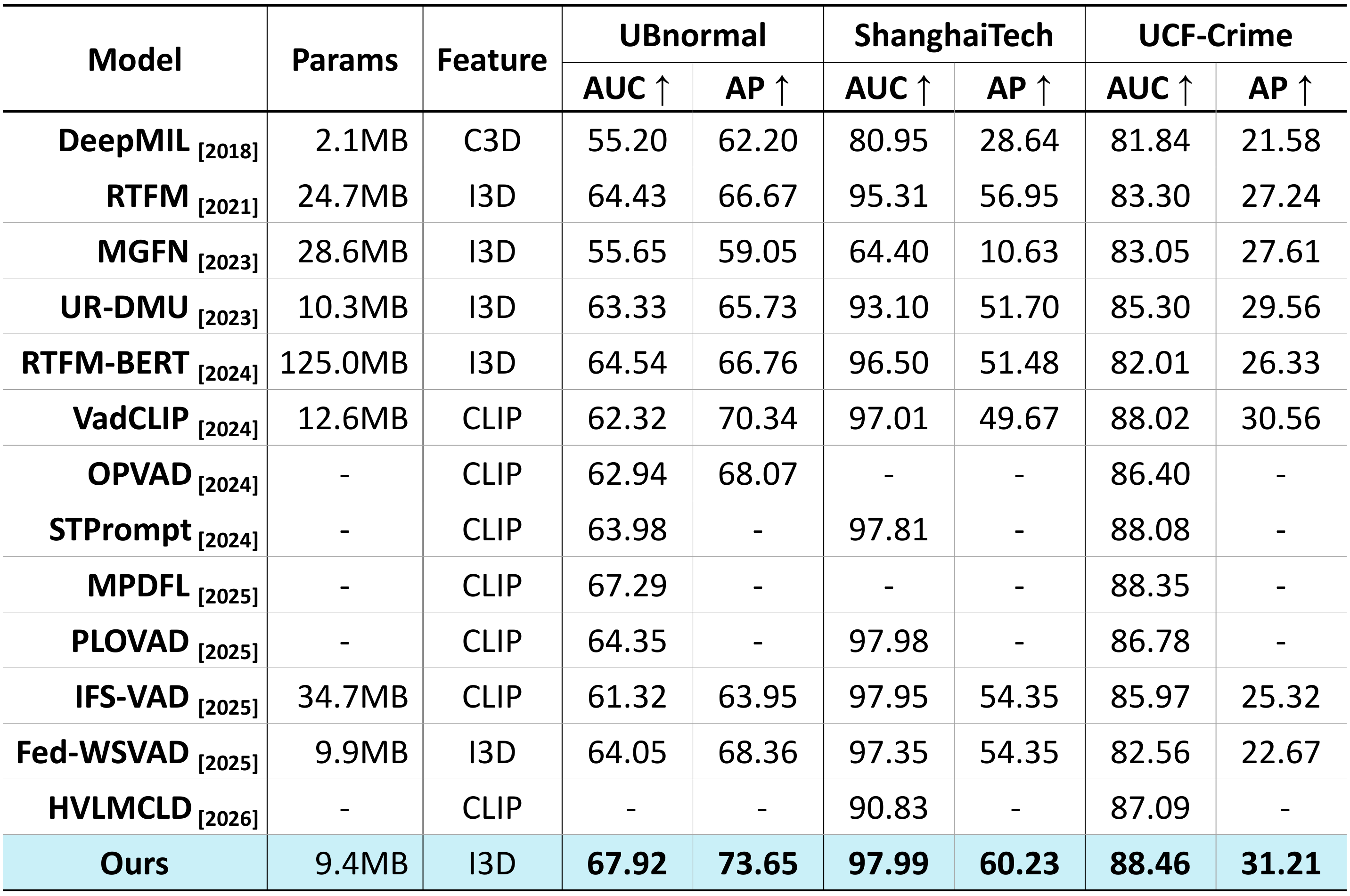}
	\end{tabular}
	\label{tab:tab1}
\end{table}

\begin{figure*}[!t]
	\centering{\includegraphics[width=1\linewidth]{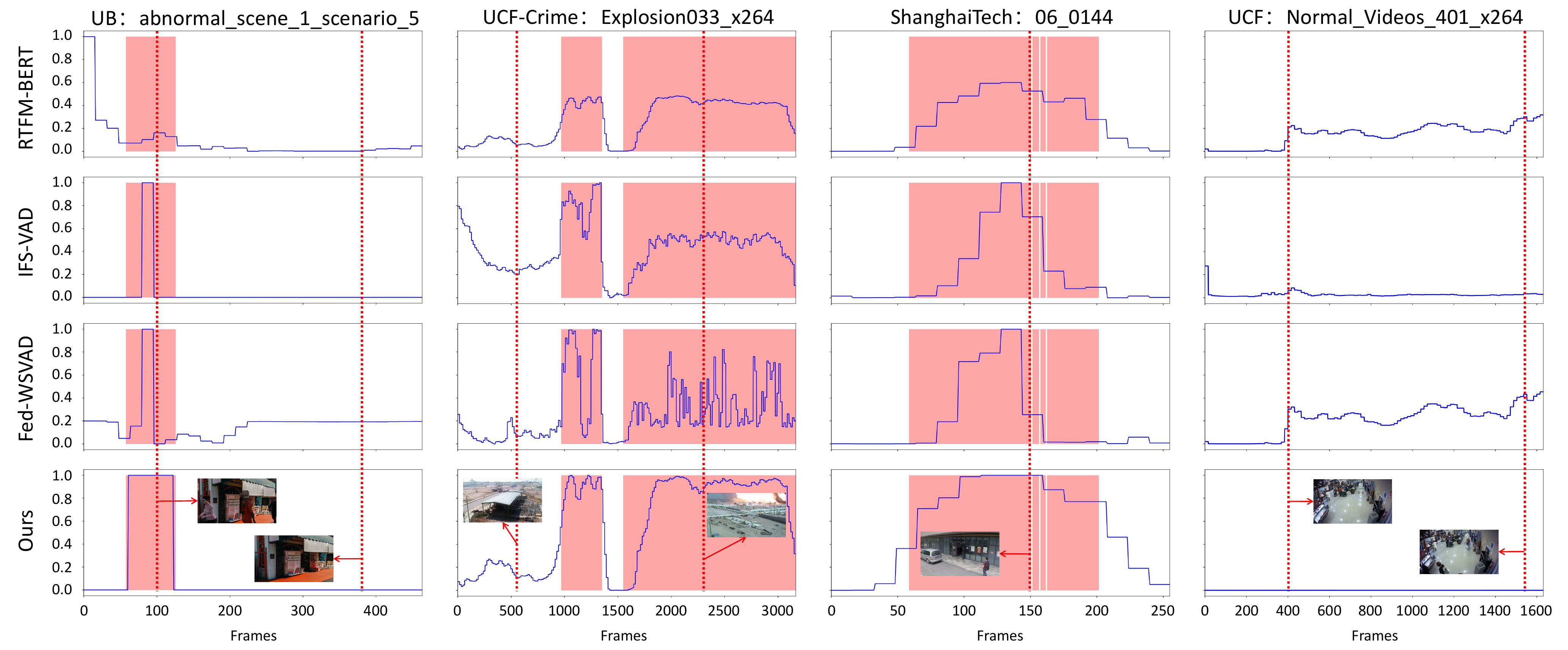}}
	\caption{The qualitative comparison of our method with RTFM-BERT, IFS-VAD and, Fed-WSVAD on the testing videos. Colored windows indicate the true abnormal regions. Enlarging the view enhances the effectiveness.}
	\label{fig:quality}
\end{figure*}

\noindent\textbf{Qualitative results.}
Figure~\ref{fig:quality} presents qualitative comparisons between our method and several representative approaches, including RTFM-BERT, IFS-VAD, and Fed-WSVAD, on videos from UBnormal, UCF-Crime, and ShanghaiTech. The red regions denote ground-truth abnormal segments. As shown in the figure, existing methods often produce unstable anomaly responses or inaccurate temporal localization. For example, RTFM-BERT tends to generate weak responses in abnormal intervals, while IFS-VAD and Fed-WSVAD exhibit noisy or fluctuating predictions across temporal frames. In contrast, our method produces anomaly scores that align closely with ground-truth abnormal regions, with clearer temporal boundaries and stable responses. Moreover, on normal videos, our method maintains low anomaly scores, reducing false alarms compared with other methods. These results demonstrate that structured evidence selection and the dual-prototype geometric decision mechanism can highlight anomaly-relevant semantics while suppressing irrelevant scene cues, leading to accurate and robust anomaly localization.

\begin{table}[!t]
	\centering
	\caption{Ablation study of the proposed framework on UCF-Crime. SAEC, RCEV, and DPGD denote the three main components, progressively enabled to evaluate their contributions.}
	\begin{tabular}{c}
		\includegraphics[width=1\linewidth]{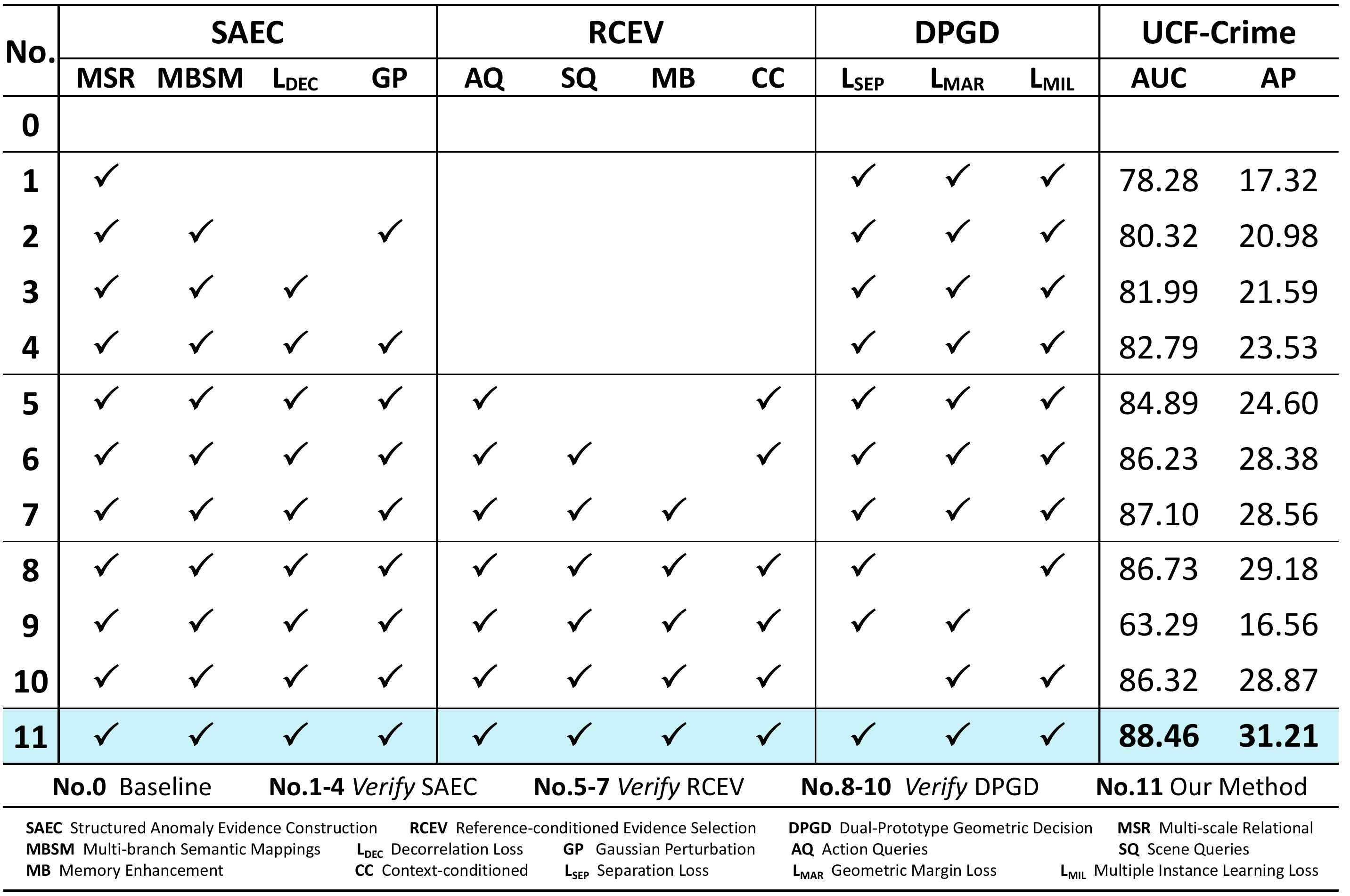}
	\end{tabular}
	\label{tab:tab2}
\end{table}

\subsection{Ablation study.}
Table~\ref{tab:tab2} presents the ablation study of the framework on the UCF-Crime dataset to analyze contributions of different components.

\noindent\textbf{Effect of SAEC.}
Rows No.1--4 evaluate the Structured Anomaly Evidence Construction (SAEC) module. Starting from the basic relational modeling (MSR), introducing multi-branch semantic mappings (MBSM) and decorrelation loss ($L_{\text{DEC}}$) progressively improves the detection performance, indicating that multi-perspective semantic representations provide richer anomaly evidence. Further incorporating Gaussian perturbation (GP) enhances the robustness of feature representations, leading to additional performance gains.

\noindent\textbf{Effect of RCEV.}
Rows No.5--7 evaluate the Reference-conditioned Evidence Selection (RCEV) module. Introducing action queries (AQ) and scene queries (SQ) enables the model to leverage contextual cues for anomaly reasoning. The memory bank (MB) further refines the query representations, while the context-conditioned aggregation (CC) adaptively selects anomaly-relevant semantic features. These components collectively improve anomaly discrimination by suppressing irrelevant scene responses.

\begin{table}[!t]
	\centering
	\caption{Impact of hyperparameters on anomaly detection performance. (A) Effect of memory bank size. (B) Effect of number of prototypes in the geometric decision module.}
	\begin{tabular}{c}
		\includegraphics[width=1\linewidth]{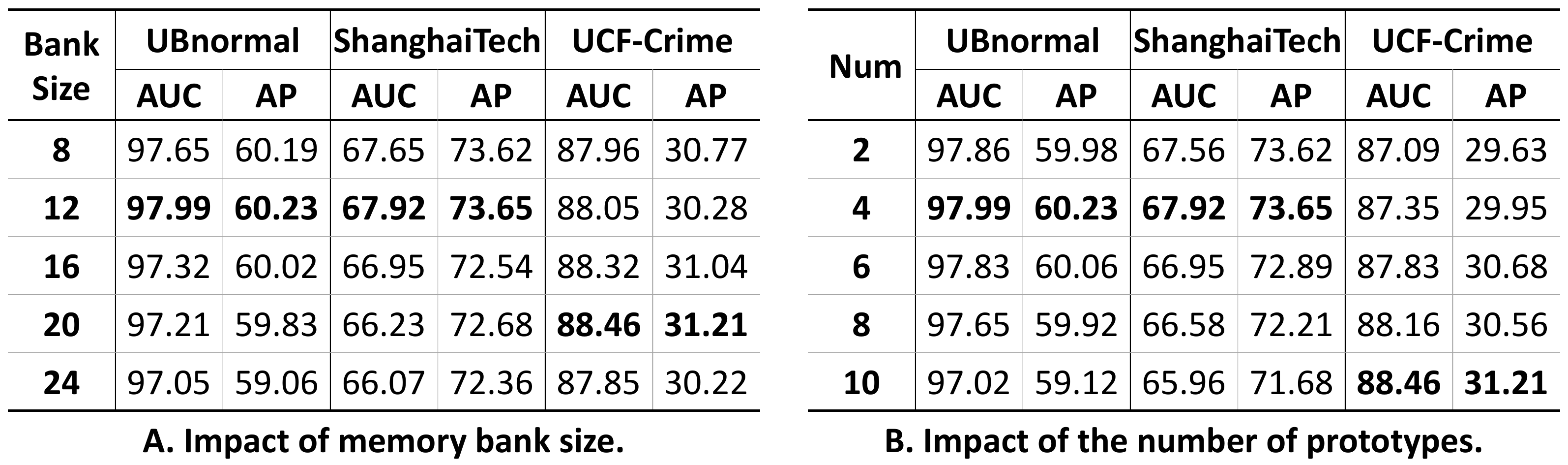}
	\end{tabular}
	\label{tab:tab3}
\end{table}

\begin{figure}[!t]
	\centering
	\begin{tabular}{c}
		\includegraphics[width=1\linewidth]{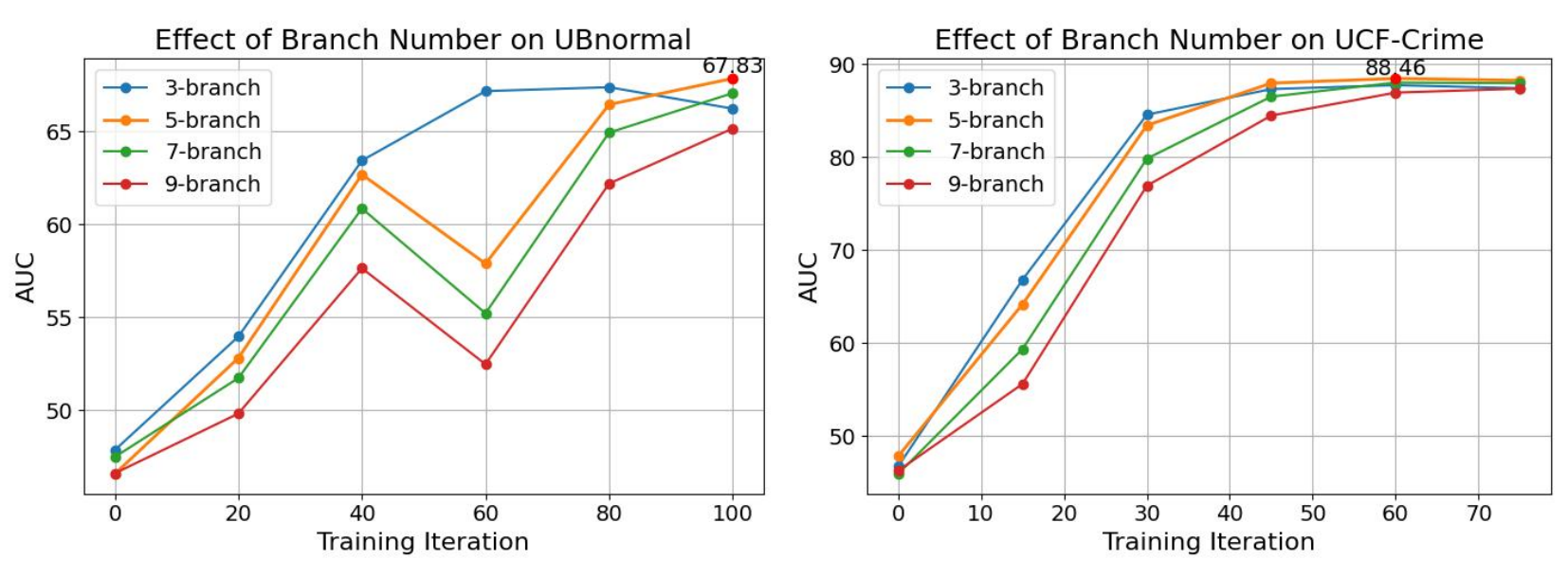}
	\end{tabular}
	\caption{Effect of branch number on anomaly detection performance over training iterations on UBnormal (left) and UCF-Crime (right). The best are obtained with 5 branches.}
	\label{fig:fig5}
\end{figure}

\noindent\textbf{Effect of DPGD.}
Rows No.8--10 analyze the Dual-Prototype Geometric Decision (DPGD) module. Removing individual loss terms leads to noticeable performance degradation, indicating that the geometric structure plays a crucial role in stabilizing anomaly discrimination. In particular, the separation loss ($L_{\text{SEP}}$) maintains sufficient distance between normal and abnormal prototypes, while the geometric margin loss ($L_{\text{MAR}}$) and MIL loss ($L_{\text{MIL}}$) jointly guide the embedding space toward a discriminative structure.

Finally, full model (No.11) integrates components and achieves performance, demonstrating effectiveness and complementarity.

\subsection{Parameter Sensitivity Analysis}
We further analyze the sensitivity of hyperparameters in the framework, including memory bank size, number of prototypes in the geometric decision module, and number of branches.

\noindent\textbf{Memory bank size.}
Table~\ref{tab:tab3} (A) shows the influence of memory bank size on anomaly detection performance. The results indicate that the optimal bank size varies across datasets. Specifically, the best results on UBnormal and ShanghaiTech are achieved when the bank size is set to 12, while UCF-Crime reaches its best performance at 20. This suggests that the required memory capacity is related to dataset complexity and semantic diversity. Overall, moderate bank sizes consistently yield strong performance, whereas overly small or overly large settings tend to be less effective.

\noindent\textbf{Number of prototypes.}
Table~\ref{tab:tab3} (B) presents the effect of the number of prototypes in the geometric decision module. A similar trend can be observed: the best performance on UBnormal and ShanghaiTech is obtained with 4 prototypes, while UCF-Crime achieves the highest performance with 10 prototypes. These results indicate that the suitable prototype number is also dataset-dependent. In general, a moderate number of prototypes is sufficient to provide effective geometric discrimination, whereas too many prototypes may introduce unnecessary complexity.

\noindent\textbf{Number of semantic branches.}
Figure~\ref{fig:fig5} illustrates the effect of the number of branches during training. The curves show that using five branches leads to stable optimization behavior and the best performance. With fewer branches, the model may not capture sufficient semantic diversity, while too many branches introduce redundant features and degrade training stability.

Overall, these experiments demonstrate that the proposed framework is relatively robust to hyperparameter variations and achieves optimal performance under moderate configurations.

\begin{table}[!t]
	\centering
	\caption{Cross-dataset generalization between ShanghaiTech and UCF-Crime. Models are trained on one and evaluated on the other. Performance is reported in AUC and AP.}
	\begin{tabular}{c}
		\includegraphics[width=1\linewidth]{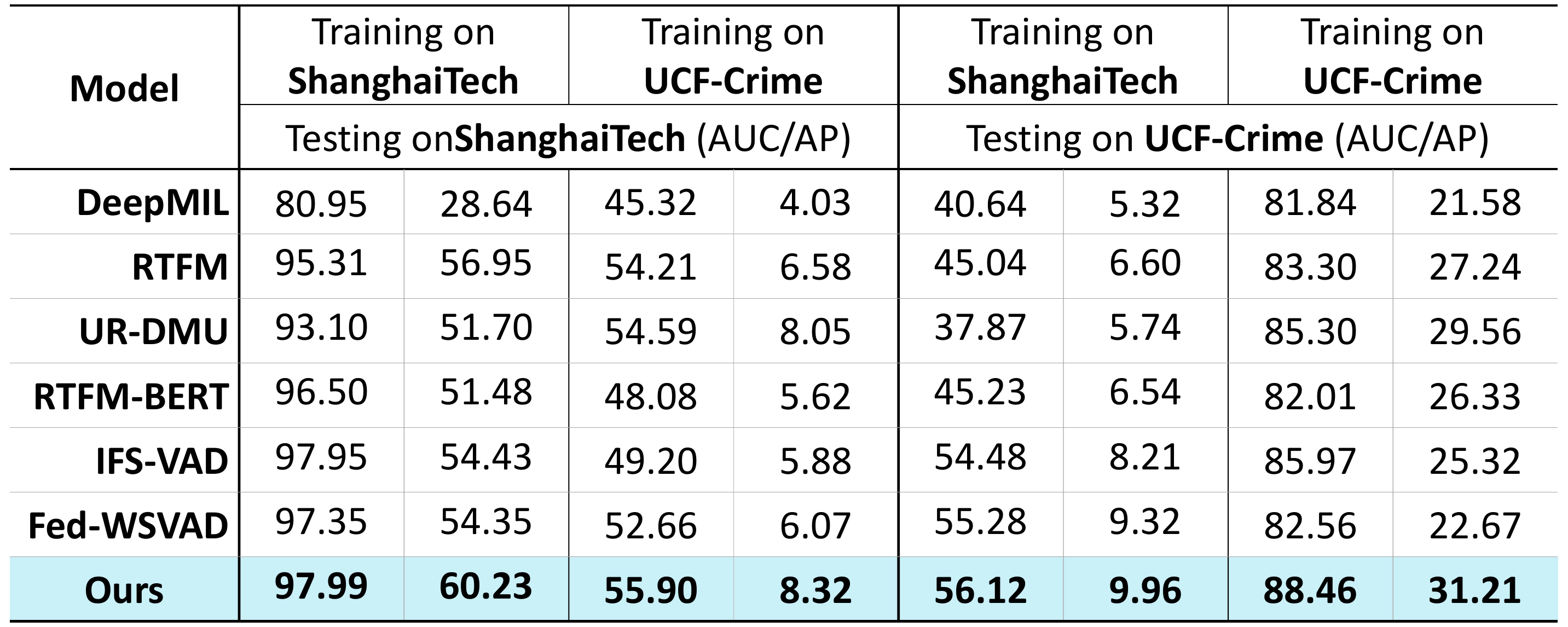}
	\end{tabular}
	\label{tab:tab5}
\end{table}

\subsection{Cross-dataset Generalization}
Table~\ref{tab:tab5} reports cross-dataset generalization results between ShanghaiTech and UCF-Crime, where models are trained on one dataset and evaluated on the other. This setting evaluates the robustness of anomaly representations under scene distribution shifts.

When trained on ShanghaiTech and tested on UCF-Crime, the proposed method achieves 56.12 AUC and 9.96 AP, outperforming all compared approaches, including representative baselines such as RTFM, UR-DMU, and IFS-VAD. Similarly, when trained on UCF-Crime and evaluated on ShanghaiTech, our method again obtains the best performance with 55.90 AUC and 8.32 AP, demonstrating consistent advantages under reverse transfer settings. These results indicate that the proposed structured evidence selection mechanism together with the dual-prototype geometric decision module enables the model to focus on anomaly-relevant semantics rather than scene-specific statistics, thereby improving robustness and cross-dataset generalization ability.

\begin{figure}[!t]
	\centering
	\begin{tabular}{c}
		\includegraphics[width=1\linewidth]{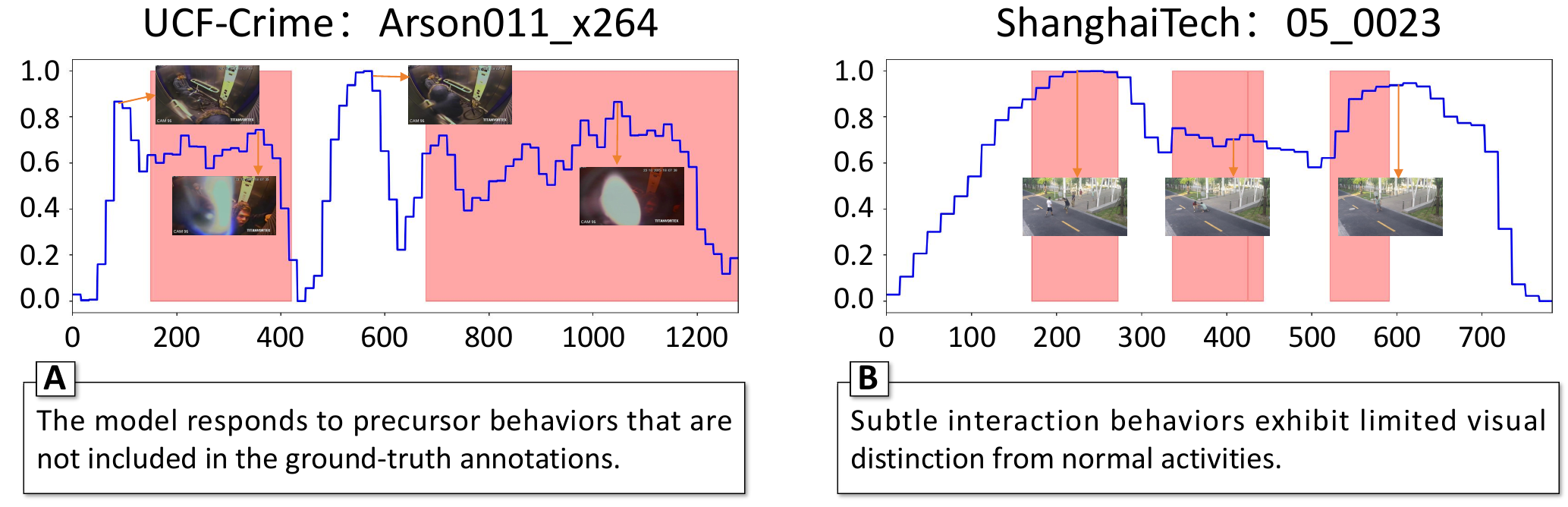}
	\end{tabular}
	\caption{Failure case analysis. (A) Precursor behaviors outside ground truth. (B) Subtle interactions with weak cues.}
	\label{fig:failure}
\end{figure}

\subsection{Failure Case Analysis}

Although the proposed framework achieves strong performance, challenging scenarios in complex real-world environments may still sometimes lead to imperfect predictions. As shown in Figure~\ref{fig:failure} (A), the model produces anomaly responses slightly earlier than the annotated abnormal segment begins. These responses correspond to potential precursor behaviors that deviate from normal activity patterns but fall outside the ground-truth annotations. Since the proposed framework searches for anomaly-related evidence through structured evidence selection, it may detect subtle early deviations that are not included in the annotation boundaries.

Figure~\ref{fig:failure} (B) illustrates a challenging scenario where anomalous interactions exhibit very weak visual distinction from normal activities. In such cases, motion patterns and scene context remain highly similar to regular behaviors, making it difficult to construct strong and discriminative anomaly evidence during evidence selection. As a result, the instance features show only limited separation from normal prototypes in the geometric decision space, leading to relatively low overall anomaly confidence prediction.

\section{Conclusion}
This paper revisits weakly supervised video anomaly detection from the perspective of structured evidence reasoning. Instead of directly regressing anomaly scores from aggregated representations, we propose a Structured Evidence Selection framework that constructs and selects anomaly-related evidence before the final decision stage. The framework organizes clip-level features into multi-perspective semantic representations, refines anomaly evidence through reference-conditioned selection with scene and action queries, and performs anomaly discrimination using a lightweight dual-prototype geometric decision mechanism. This structured reasoning process explicitly highlights anomaly-relevant semantics while suppressing scene interference, thereby improving detection robustness and stability, leading to overall more stable anomaly discrimination under weak supervision.

In future work, we plan to explore adaptive evidence construction and investigate how structured evidence reasoning can be integrated with foundation-model video representations. Extending the framework to open-world anomaly detection and more complex environments is also a promising direction.

\bibliographystyle{ACM-Reference-Format}
\bibliography{software}
\end{document}